\documentclass[10pt,twocolumn,letterpaper]{article}

\usepackage[pagenumbers]{iccv} 

\usepackage{dsfont}

\usepackage{algorithm}
\usepackage{algpseudocode}
\usepackage{colortbl} 
\usepackage{graphicx} 
\usepackage{multirow} 
\usepackage{bbm}

\usepackage{setspace}

\newcommand{\set}[1]{\mathcal{#1}}
\definecolor{iccvblue}{rgb}{0.21,0.49,0.74}
\usepackage[pagebackref,breaklinks,colorlinks,allcolors=iccvblue]{hyperref}
\def\paperID{7887} 
\def\confName{ICCV}
\def\confYear{2025}

\title{Exploiting Vision Language Model for Training-Free 3D Point Cloud OOD Detection via Graph Score Propagation}

\author{Tiankai Chen$^1$, Yushu Li$^2$$^3$, Adam Goodge$^3$, Fei Teng$^1$, Xulei Yang$^3$, Tianrui Li$^1$, Xun Xu\textsuperscript{*}$^3$\\
$^1$School of Computing and Artificial Intelligence,Southwest Jiaotong University\\
$^2$South China University of Technology\\
$^3$Institute for infocomm research, A*STAR I$^2$R\\
}

\begin{document}
\maketitle
\begin{abstract}
Out-of-distribution (OOD) detection in 3D point cloud data remains a challenge, particularly in applications where safe and robust perception is critical. While existing OOD detection methods have shown progress for 2D image data, extending these to 3D environments involves unique obstacles. This paper introduces a training-free framework that leverages Vision-Language Models (VLMs) for effective OOD detection in 3D point clouds. By constructing a graph based on class prototypes and testing data, we exploit the data manifold structure to  enhancing the effectiveness of VLMs for 3D OOD detection.  We propose a novel Graph Score Propagation (GSP) method that incorporates prompt clustering and self-training negative prompting to improve OOD scoring with VLM. Our method is also adaptable to few-shot scenarios, providing options for practical applications. We demonstrate that GSP consistently outperforms state-of-the-art methods across synthetic and real-world datasets 3D point cloud OOD detection. The code is available on \url{https://github.com/handsome999KK/GSP_OOD}
\end{abstract}    
\section{Introduction}
\label{sec:intro}

\begin{figure}[!htp]
  \centering
  \includegraphics[width=0.99\linewidth]{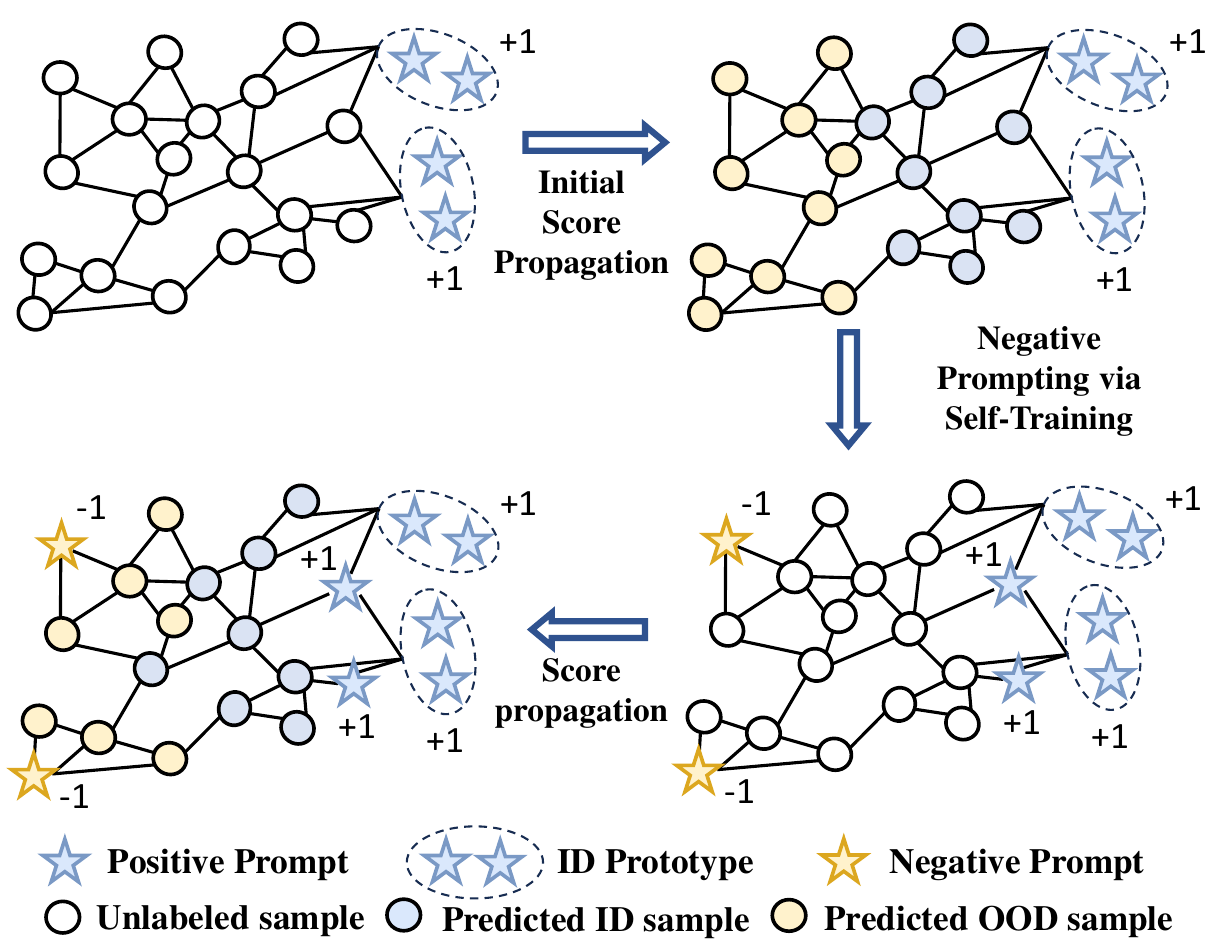}
  \vspace{-0.2cm}
  \caption{Exploiting the testing data manifold for zero-/few-shot OOD detection. ID prototypes propagate scores to test samples, with confident ones selected as pseudo negative prompts for refinement.}
  \vspace{-0.4cm}
  \label{fig:SelfTrain}
\end{figure}


The detection of out-of-distribution (OOD) samples in 3D point cloud data is critical for the safe and robust deployment of deep learning models in real-world environments. For example, identifying unexpected objects is crucial for ensuring the safety of personnel, vehicles, and robots by preventing collisions~\cite{qian20223d, nitsch2021out, paper7}. Existing approaches to OOD detection in 3D point clouds often focus on training separate models for specific tasks~\cite{paper25}. These methods typically rely on abundant in-distribution (ID) training data to build models based on reconstruction~\cite{paper24}, distance metrics~\cite{chen2021adversarial}, or discriminative strategies~\cite{paper1}. While these methods achieve strong results, they depend on substantial ID training data, which is often challenging to collect in many real-world applications and generalization may be hindered by the limited diversity of ID training samples. These limitations highlight the need for alternative solutions that do not require extensive ID data collection.

Vision-language models (VLMs) have emerged as powerful foundation models that bridge visual features with textual understanding~\cite{CLIP, li2022blip, jia2021scaling}. VLMs are trained to associate the content of visual inputs with their textual labels and they demonstrate strong zero-shot generalization to unseen visual content~\cite{CLIP}. This capacity makes VLMs appealing for OOD detection without retraining~\cite{ming2022delving, paper11, paper13, paper14}. A common VLM-based OOD detection pipeline maps ID class concepts, such as class names with various templates, into a shared feature space using the text encoder. The visual samples, similarly projected using the visual encoder, are then scored by comparing their distances to the ID class concepts in this feature space~\cite{ming2022delving}.

This VLM-based OOD detection pipeline has shown competitive performance for 2D image detection, primarily due to the vast understanding of 2D VLMs from the massive-scale image-text pre-training datasets (e.g. 400M pairs for pre-training CLIP \cite{CLIP}). In contrast, VLMs for 3D point clouds, such as the state-of-the-art ULIP2~\cite{paper26}, are typically pre-trained on much more restricted datasets (e.g., 800k point cloud samples). This restricts their ability to generalize across unseen categories and distributional shifts (e.g., style changes), which reduces the effectiveness of similarity-based methods for 3D point cloud OOD detection.

In this work, we propose to leverage the structure of the testing data manifold to compensate for this deficiency and improve VLM adaptation for downstream tasks~\cite{stojnic2024label, Li2025efficient}, as illustrated in Fig.~\ref{fig:SelfTrain}. Specifically, we construct a graph based on class prototypes (e.g., text prompts for ID classes) and testing data features, where prototype nodes are assigned labels that propagate to neighboring testing nodes. Given the limited training data, we hypothesize that exploiting the testing data manifold enhances the generalization capabilities of 3D point cloud VLMs. Our method, Graph Score Propagation (GSP), assigns a positive OOD score (e.g., $+1$) to ID class prototypes while all testing samples start with a neutral score (0). As scores propagate along the graph, the resultant scores indicate OOD likelihood.

This straightforward adoption of score propagation significantly improves upon existing distance-based methods, and we introduce two additional enhancements to further boost OOD detection accuracy. First, while conventional training-free approaches typically use a single class prototype by averaging text prompts~\cite{stojnic2024label,zhang2022tip}, we argue that capturing the prompt distribution through clustering better leverages prompt diversity. Second, negative prompting has shown strong performance in VLM-based OOD detection~\cite{tian2024argue,paper10,paper12}. However, simply negating the description (e.g. add ``not'') is not very effective and learning negative prompts requires sophisticated optimization. Instead, we employ self-training to convert high-confidence OOD testing samples into pseudo-negative prompts and assign them a negative score (e.g., $-1$), for subsequent propagation iterations. Finally, our framework integrates seamlessly into traditional supervised OOD detection protocols by incorporating a few ID samples as positive prompts, further enhancing OOD detection under both zero-shot and few-shot protocols.

Our contributions are mainly summarized as follows.

\begin{itemize}
    \item We employ a pre-trained vision-language model for 3D point cloud OOD detection and introduce Graph Score Propagation (GSP), a novel methodology that leverages the testing data manifold instead of solely relying on text-test feature distances.
    \item We enhance GSP with prompt clustering and self-training-based negative prompting. GSP is also compatible with both zero-shot and few-shot testing protocols.
    \item We validate the effectiveness of our method on synthetic and real-world 3D point cloud OOD detection benchmarks, consistently achieving superior performance over state-of-the-art VLM-based OOD detection methods.
\end{itemize}

\section{Related Work}

\noindent\textbf{Out-of-Distribution Detection}: Traditional OOD detection methods focused on softmax-based probability scoring \cite{paper1}. Recently, attention has shifted to novel evaluation metrics \cite{paper2,paper3,paper4,paper5}, such as differential network channel processing \cite{paper4,paper5} and non-parametric nearest-neighbor distances \cite{paper2}.  Other strategies involve training specific OOD discriminators, like DAL \cite{paper6}, which constructs an OOD distribution set, and MOOD \cite{paper8}, which uses masked image modeling. NECO \cite{paper9} utilizes neural collapse geometry for OOD identification. Vision-language models have introduced new methods using text prompts to distinguish between in-distribution (ID) and OOD data \cite{paper10,paper11,paper12,paper13,paper14}. LoCoOP \cite{paper14} leverages CLIP's local features for OOD regularization, while Bai et al. \cite{paper11} use CLIP to detect ID-like outliers, enhancing OOD detection and few-shot learning. Training text encoders \cite{paper13} and learnable prompt \cite{paper10,paper12} further increases the score gap between ID and OOD data. Recently, research has expanded to include 3D datasets for point cloud OOD detection \cite{paper15,paper16}. Studies have shown that 3D OOD detection is particular important for autonomous driving scenarios~\cite{paper7, Ksel2024RevisitingOD}.

\noindent\textbf{3D Vision-Language Models}: Recent advancements in Vision-Language Models (VLMs) have facilitated innovative approaches for 3D understanding. Techniques utilizing CLIP \cite{paper19,paper20,paper21} transform 3D point clouds into imagery data for extracting features aligned with text descriptions. More recently, methods like ULIP \cite{paper25,paper26} learn unified representations across images, text, and 3D point clouds, further improving zero-shot 3D classification. These approaches leverage the integration of text prompts and visual features which could better distinguish between in-distribution and out-of-distribution data in 3D contexts.



\noindent\textbf{Label Propagation}: Label propagation (LP) is a pivotal method in semi-supervised learning, leveraging limited labeled data to infer labels for unlabeled instances by exploiting data similarity across a graph \cite{iscen2019label,paper30,paper31,paper32}. It assumes that labels vary smoothly across the graph, making it effective in transductive learning settings \cite{zhu2003semi, iscen2019label}. Recent advancements integrate LP with graph neural networks to enhance its ability to separate ID from OOD data \cite{Wu2023EnergybasedOD, paper33,paper34}. In the context of VLMs, LP employs text prompts as label representations, constructing graphs that propagate labels from text prototypes to test samples, thus improving zero-shot 2D classification \cite{hu2024reclip, stojnic2024label,Li2025efficient}. Contrary to the existing attempts of combining LP with VLM, we tackle a one-class classification task. Only prototypes for in-distribution classes are available. We further propose a novel negative prompting technique via self-training and prompt clustering technique to better separate ID from OOD.
\section{Methodology}
\label{sec:intro}

\begin{figure*}[!htb]
  \centering
  \includegraphics[width=0.99\linewidth]{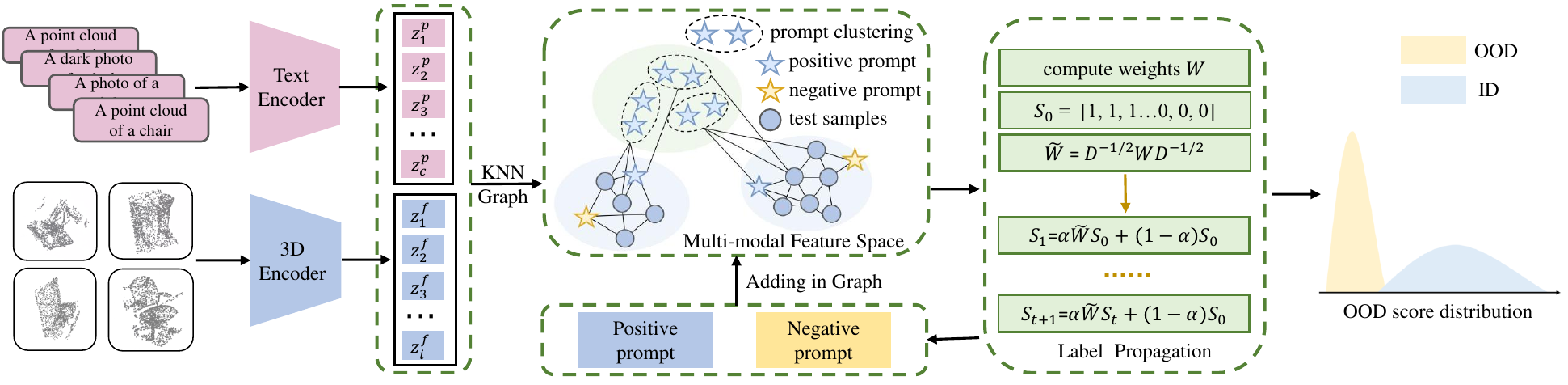}
\vspace{-0.2cm}
  \caption{Overview of the proposed framework for zero-/few-shot OOD detection via graph score propagation. The framework encodes ID category prompts and dataset samples using frozen text and 3D encoders, constructing a KNN graph. Initial score propagation generates pseudo prompts, which refine the graph before final score propagation computes OOD scores.}
  \vspace{-0.3cm}
  \label{figure1}
\end{figure*}

\subsection{Problem Formulation}

We begin by formally defining the task of VLM-based out-of-distribution (OOD) detection in 3D point clouds. Let  \( z^f = f(x; \theta) \) and \( z^t = g(t; \phi) \) denote the encoded features of point cloud data $x$ and text $t$, where $f(\cdot ; \theta)$ and $g(\cdot; \phi)$ are the VLM visual and text encoders respectively. The task involves the encoded features of unlabeled test data \( \mathcal{D}_u = \{f(x_i)\}_{i=1\cdots N_u} \) and, optionally, the features and labels of few-shot labeled data \( \mathcal{D}_l = \{f(x_j), y_j\}_{j=1\cdots N_l} \). The set of in-distribution (ID) classes \( \mathcal{Y}_{in} = \{1, \cdots, C_{in}\} \) is assumed to be known. Following established practices for VLM adaptation~\cite{ming2022delving}, we represent each ID class with $N_t$ prompt templates \( \mathcal{T} = \{t_{j}(y)\}_{N_t} \), e.g., ``This is a point cloud of \{y\}.'' or ``A good point cloud of \{y\}''. These textual prompts are encoded by the text encoder, creating prototypes for ID classes \( \mathcal{P} \). Existing works often create prototypes by averaging the encoded prompts for each class, resulting in a prototype \( z^p_c \) for the \( c \)-th class, and the full set of prototypes is denoted \( \mathcal{P} = \{z^p_c\}_{c=1\cdots C_{in}} \).
The OOD detection task is to determine whether a test sample \( x_i \in \mathcal{D}_u \) belongs to an ID class, leveraging ID prototypes \( \mathcal{P} \) and, optionally, the few-shot labeled data \( \mathcal{D}_l \).

\noindent\textbf{Preliminary of VLM}:
We briefly review the training objective of vision-language models (VLMs). The pre-training steps align the visual features with their associated text features~\cite{paper20, paper25, paper26}, typically via a contrastive loss function:

\begin{equation}
\mathcal{L}_{vlm} = \sum_i \log \frac{\exp(z^f_i z^t_i / \tau)}{\sum_j \exp(z^f_i z^t_j / \tau)}
\end{equation}

When a VLM is trained on a large amount of point cloud-text pairs, the cosine similarity of positive pairs (a point cloud sample and text prompt of its ground-truth class) is higher than that of negative pairs. Therefore, a pre-trained VLM provides a straightforward solution to 3D point cloud OOD detection by measuring the cosine similarity between textual descriptions of in-distribution classes and testing sample features. A high distance in the visual-text feature space suggests a greater likelihood of being OOD.

\subsection{VLM for OOD Detection}

\noindent\textbf{Vanilla OOD Detection with VLM}: 
Pre-trained VLMs offer an effective approach to OOD detection by measuring the distance between visual samples and text labels in the feature space. For example, \cite{ming2022delving} employs the maximum normalized cosine similarity as the OOD score. We define a normalized score \( S \) to indicate OOD-ness, where \( S \to 0 \) suggests OOD and \( S \to 1 \) suggests ID.

\begin{equation}
    S = \max_c \frac{\exp(-d(z^f, z_c^p) / \tau)}{\sum_{j} \exp(-d(z^f, z_j^p) / \tau)}
\end{equation}

It is assumed that in the feature space, OOD samples are further from the ID prototypes than ID samples. Subsequent research has further explored this approach by introducing techniques like negative prompts to better separate OOD from ID samples~\cite{jiang2024negative}. In this work, we are motivated by the challenge of measuring similarity/distance in the VLM feature space and propose to tackle the VLM OOD detection problem by exploiting the manifold of testing and, optionally, few-shot labeled data.

\noindent\textbf{Manifold Distance for OOD Detection}: 
The vanilla measurement overlooks the complex manifold of testing data. 
A straightforward approach to consider testing data manifold for OOD detection is to replace the standard distance metric with a manifold distance, represented via a graph.

Given a graph representing the data manifold \( G = (E, V) \) with weight matrix \( W \in \mathbb{R}^{(|\mathcal{P}| + |\mathcal{D}_l| + |\mathcal{D}_u|) \times (|\mathcal{P}| + |\mathcal{D}_l| + |\mathcal{D}_u|)} \), the manifold distance between two nodes on the graph is defined as the shortest path. The shortest path between arbitrary nodes can be obtained via Dijkstra's algorithm~\cite{dijkstra2022note}. More concretely, this can be formulated as finding a \textit{path} between a prototype \( z^p\in\set{P} \) or a few-shot labeled sample \( z_k^f\in\set{D}_l \) and a testing sample \( z_j^f\in\set{D}_u  \), consisting of edges \( \tilde{E}_{ij} = \{e_k\} \) on the graph, such that the sum of these edge weights is minimized. The manifold distance between two nodes is thus defined as $d(z_j, z_i) = \min_{\tilde{E}_{ij}} \sum_{e_k \in \tilde{E}_{ij}} d(e_k)$, where \( d(e_k) \) denotes the distance for edge \( e_k \), e.g. the L2 distance between the encoded features.

Our empirical results in Tab.~\ref{tab:ablation} reveal that incorporating the manifold improves the effectiveness of OOD detection. However, there are two main disadvantages associated with distance-based OOD detection. First, the manifold distance is unnormalised, if directly using the reciprocal of manifold distance as OOD score the result may be biased by the relative scale. Additionally, it is non-trivial to integrate negative prompts within the distance-based OOD detection paradigm.





\subsection{VLM OOD Detection as Score Propagation}\label{sect:OODLP}

Instead of directly measuring the distance for OOD detection, we formulate OOD detection as a score propagation problem. Specifically, we assign a positive score (+1) to ID prompts $\set{P}$, while all testing samples are initially assigned a neutral score (0). Labels are then propagated from the positive nodes to all testing samples in an iterative manner. We follow a standard propagation rule~\cite{zhu2002learning} as shown in Eq.~\ref{eq:LP}, 
\begin{equation}\label{eq:LP}
\vspace{-0.2cm}
\begin{split}
    &S_t = \tilde{W} S_{t-1} + \alpha S_0,\\
    \text{s.t.} \quad &\tilde{W} = D^{-\frac{1}{2}} W D^{-\frac{1}{2}}, \; D = \textbf{diag}(\sum_j W_{ij}),\\
    &S_0 = [ \underbrace{1, \cdots, 1}_{|\mathcal{P}| + |\mathcal{D}_l| \text{ elements}}, \underbrace{0, \cdots, 0}_{|\mathcal{D}_u| \text{ elements}} ],
\end{split}
\end{equation}
where each step of propagation updates the score for unlabeled testing samples as a weighted average of scores from adjacent nodes and initial scores \( S_0 \) until $T$ iterations are completed. Consequently, nodes that are closer to the prototypes on the manifold will obtain a higher score.

\noindent\textbf{Prompt Clustering}: 
The above propagation strategy propagates labels from ID prototypes to testing samples on the graph. Thus, it is critical to design effective ID prototypes that capture the diversity of textual descriptions for ID classes. Contrary to existing practices that average the text prompt features across all prompt templates, we propose modeling the distribution of prompt templates with multiple representative ones. 
Given the prompt templates \( \mathcal{T} \), we first initialize a pool of text prompts \( \mathcal{P}_c = \{z^p_{cj} = g(t_j(y_c))\}_{j=1\cdots |\mathcal{T}|} \) for the \( c \)-th class. We then introduce a clustering step to reduce this set to \( N_c \) representative prototypes \( \tilde{\mathcal{P}}_c \). This is achieved by applying K-means \cite{2023K} clustering on \( \mathcal{P}_c \), and the cluster centers are adopted as the ID prototypes:

\begin{equation}\label{eq:prompt_cluster}
\vspace{-0.3cm}
\begin{split}
    &\tilde{\mathcal{P}}_c = \arg\min_{\{\mu_k\}, \gamma} \sum_{N_c} \sum_{z^p_{cj} \in \mathcal{P}_c} \gamma_{jk} \| z^p_{cj} - \mu_k \|^2, \\
    & \text{s.t.} \quad \gamma \in \{0, 1\}^{|\mathcal{T}| \times N_c}, \; \sum_k \gamma_{jk} = 1,
\end{split}
\end{equation}

where \( \mu_k \) are the cluster centers, and \( \gamma_{jk} \) indicates the assignment of each prompt feature \( z^p_{cj} \) to its closest center. The new prototypes, $\tilde{\set{P}}=\tilde{\set{P}}_1\cup\tilde{\set{P}}_2\cup\cdots\tilde{\set{P}}_C$, are used for score propagation as positive nodes.





\noindent\textbf{Negative Prompting via Self-Training}: 
Negative prompts have been shown to substantially benefit OOD detection~\cite{tian2024argue, paper10, paper12}. Incorporating negative prompts is particularly straightforward within the graph label propagation framework. We treat negative prompts as nodes \( \mathcal{P}^n \) with a label of \( -1 \) on the graph. During propagation, the negative labels compete with the positive labels (ID prompts) to influence the score of testing samples.
Existing approach exploits ID data to learn negative prompt~\cite{tian2024argue} which prohibits zero-shot inference.
To address this, we propose a strategy that generates negative prompts via self-training, as illustrated in Fig.~\ref{fig:SelfTrain}. Specifically, we first conduct initial score propagation for $T$ iterations following Eq.~\ref{eq:LP}, giving rise to initial propagated scores ${S}_T$. We select the top $m\%$ most confident scores from ${S}_T$ following Eq.~\ref{eq:topm_idx} where $\text{top}(S,m\%)$ indicates the top $m\%$ value of $S$.
\begin{equation}\label{eq:topm_idx}
\vspace{-0.2cm}
\begin{split}
    &\set{I}_{pos} = \{i \;|\; {S}_T[i] > \text{top}({S}_T,m\%)\},\\
    &\set{I}_{neg} = \{i \;|\; {S}_T[i] < \text{top}({S}_T,1-m\%)\}
\end{split}
\end{equation}

We label the selected samples as positive and negative prompts respectively and re-initialize the score $\tilde{S}_0$. For negative prompts, we assign scores of $-1$ to further increase the gap between ID and OOD samples. The final score is obtained by running Eq.~\ref{eq:LP} with $\tilde{S}_0$ as initial scores.

\begin{equation}\label{eq:neg_prompt}
\tilde{S}_0[i]=
\begin{cases}
    S_0[i] & \text{if } i \notin \set{I}_{pos} \;\&\; i \notin \set{I}_{neg}, \\
    +1 & \text{if } i \in \set{I}_{pos}, \\
    -1 & \text{if } i \in \set{I}_{neg}.
\end{cases}
\end{equation}

\noindent\textbf{Graph Construction}:
We now introduce the approach for constructing a graph for score propagation. As observed by \cite{Li2025efficient, stojnic2024label}, the distance between text prompts and visual samples is generally larger than the distance between visual samples. This observation also holds for 3D point cloud VLM because the 3D VLMs are often finetuned from 2D VLMs~\cite{paper25,paper26}. Therefore, we adopt a similar approach to construct the graph. We write the weight matrix \( W \) in a blockwise form as shown in Eq.~\ref{eq:adj_block}, where \( W_u \in \mathbb{R}^{|\mathcal{D}_u| \times |\mathcal{D}_u|} \), \( W_{up} \in \mathbb{R}^{|\mathcal{D}_u| \times |\tilde{\mathcal{P}}|} \) and \( W_{ul} \in \mathbb{R}^{|\mathcal{D}_u| \times |\mathcal{D}_l|} \). No connections are created between prototypes and few-shot labeled samples. All blocks are further constructed as a k-nearest neighbor (KNN) graph following~\cite{Li2025efficient, stojnic2024label}. It is also computationally efficient to adapt an existing graph to new testing samples by incrementally adding the testing samples to the graph~\cite{Li2025efficient}.

\begin{equation}\label{eq:adj_block}
\vspace{-0.1cm}
     W = \left[
    \begin{array}{ccc}
        \mathbf{I}^{|\tilde{\set{P}}|\times |\tilde{\set{P}}|} & \mathbf{0} & W_{up}^\top \\
        \mathbf{0} & \mathbf{I}^{|\set{D}_l|\times |\set{D}_l|} & W_{ul}^\top \\
        W_{up} & W_{ul} & W_{u}
    \end{array}
    \right]
\end{equation}

\noindent\textbf{Overall Algorithm}:
We present the overall algorithm for unified iterative label propagation for vision-language models in Algorithm \ref{alg:example}. In Fig.~\ref{figure1}, we show an overview of the proposed framework for zero-/few-shot OOD detection via graph propagation.

\begin{algorithm}
\caption{Graph Score Propagation for OOD Detection}
\label{alg:example}
\begin{spacing}{1.2}
\begin{algorithmic}[1]
    \State \textbf{Input:} training data $\set{D}_l$, testing data $\set{D}_u$ and ID class prompt templates $\set{T}_c$
     \State \textbf{Output:} Final OOD scores $\tilde{S}_T$
    \State \textbf{for} $\set{D}_l$ and  $\set{T}_c$, \textbf{do}

Encode: $z_l = f(x_l)$ for training data and $\set{P}_c = \{g(t_j(y_c))\}$ for prompt templates.

Prompt clustering: 
      positive prototypes $\tilde{\set{P}}$ \hfill $(\text{Eq.~\ref{eq:prompt_cluster}})$

\State \textbf{end for}    
    
    \State \textbf{for} $\set{D}_u$, \textbf{do}
    
    Encode: 
      $z_i = f(x_i)$ for testing data 
      
    Build graph to compute $W$ 
      \hspace*{\fill} $(\text{Eq.~\ref{eq:adj_block}})$
      
    Initialize score $S_0^{(i)}$ 
      \hfill $(\text{Eq.~\ref{eq:LP}})$
    
   Score propagate to compute $S_T$\hfill $(\text{Eq.~\ref{eq:LP}})$
    
  Self-training to generate positive $\set{I}_{pos}$ and negative prompts $\set{I}_{neg}$ \hfill $(\text{Eq.~\ref{eq:topm_idx}})$
    
Updated initial scores $\tilde{S}_0$ \hfill $(\text{Eq.~\ref{eq:neg_prompt}})$

Score propagate to compute $\tilde{S}_T$ \hfill $(\text{Eq.~\ref{eq:LP}})$
  \State \textbf{end for}    
  \State \textbf{return} $\tilde{S}_T$   
\end{algorithmic}
\end{spacing}
\end{algorithm}









\vspace{-0.2cm}

\section{Experiments}
\subsection{Experimental setup}

\noindent\textbf{Dataset}: In this work, we follow the 3DOS framework \cite{paper15} to construct downstream datasets for evaluating our method. The 3DOS benchmark is based on two well-known 3D object datasets, \textbf{ShapeNetCore}~\cite{paper35} and \textbf{ScanObjectNN}~\cite{paper37}. 
ShapeNetCore includes 
54 categories covering daily objects. ScanObjectNN is cropped out from ScanNet dataset featuring a real-world objects. It consists of 15 categories with 2,902 3D objects. 
To further validate our approach on real-world data, we introduce the \textbf{Sydney Urban Object Dataset}~\cite{de2013unsupervised}, \textbf{nuScenes}~\cite{nuScenes} and \textbf{S3DIS}~\cite{S3DIS}. Sydney Urban Object Dataset consists of LiDAR-scanned point clouds across 26 categories, totaling 631 unique instances. S3DIS dataset contains 6 large-scale indoor areas with 271 rooms annotated into 13 semantic categories. We cropped out individual objects using ground-truth segmentation masks, leading to 8931 objects. The nuScenes Dataset features point clouds captured by LiDAR sensor under autonomous driving scenario. We  cropped out point clouds using ground-truth bounding boxes, leading to 2205 objects across 9 categories. Detailed visualization of each dataset is deferred to the supplementary.

\noindent\textbf{Evaluation Protocol}: Following the methodology outlined in 3DOS \cite{paper15}, we divide ShapeNetCore into three non-overlapping category sets, SN1, SN2, and SN3, referred to as the ShapeNetCore dataset. Each set contains 18 categories. For each OOD test scenario, one of these sets (SN1, SN2, or SN3) is treated as the known set of categories, while the remaining two sets are used as unknown categories, resulting in three distinct OOD test cases. Similarly, 3DOS defines three category sets, SR1, SR2, and SR3 on the ScanObjectNN dataset. Each SR-set includes 5 non-overlapping categories, with the other two sets acting as unknown categories in each scenario and call them as ScanObjNN. As for Sydney Urban Object Dataset, we categorize these objects into two groups: movable objects, e.g. ``truck'' ``pedestrian'' and non-movable objects, e.g. ``building'' and ``traffic sign''. Movable objects are treated as in-distribution (ID) classes, simulating a scenario in autonomous driving where object detection primarily focuses on movable objects. For consistency, we pre-process each point cloud sample by upsampling it to a uniform number of points. For S3DIS dataset, We retained instances containing more than 2048 original points, where each instance was randomly downsampled to retain exactly 2048 points. Then we set the fore-ground objects as ID and back-ground objects as OOD. For nuScences dataset, we have conducted preliminary screening on the original point cloud and retained instances with more than 200 original points and upsampling it to 1024 points. Then we define Four-wheeled vehicles~(including car, bus, truck, construction\_vehicle, trailer) as ID and the rest as OOD. 
And further details provided in the supplementary material.



\noindent\textbf{Implementation Details}. We use the pre-trained ULIP/ULIP2 \cite{paper25,paper26} model to extract point cloud features and text features. For the 3D encoder, we select Point-BERT \cite{paper38}. In constructing the KNN graph, we set a universal K-value(\( N_k\)) of 10, connecting each text feature to its 10 nearest point cloud features. In Eq.~\ref{eq:LP}, we set $\alpha$ to 0.5 and the number of iterations $T$ to 5. The percentage of pseudo negative prompts, i.e. $m\%$ in Sect.~\ref{sect:OODLP}, is also set to 5\%. Additionally, we use K-means clustering(\( N_c\)) to group the prompts for each category into 3 clusters. 


\noindent\textbf{Competing Methods}. We compare our method with several OOD detection approaches, including full-shot methods from 3DOS \cite{paper15}. We also benchmark against recent zero-shot methods, such as DDCS \cite{paper5}, ZLaP\cite{stojnic2024label},
MCM \cite{ming2022delving}, and NegLabel \cite{jiang2024negative}. For zero-shot comparisons, all methods use the 3D encoder from \cite{paper26}. Baseline results are also reported by directly using cosine similarity as the OOD score, with these methods referred to as ULIP and ULIP2, depending on the VLM used.

\begin{table*}[htbp]
  \centering
  \setlength{\tabcolsep}{1.3pt} 
  \renewcommand{\arraystretch}{0.9} 
  \fontsize{8}{9}\selectfont
  \resizebox{\textwidth}{!}{
    \begin{tabular}{l|l|cc|cc|cc|cc|cc|cc|cc|cc}
    
       \hline
      \toprule
       & & \multicolumn{8}{c|}{\textbf{ScanObjNN}} & \multicolumn{8}{c}{\textbf{ShapeNetCore}}\\
      & & \multicolumn{2}{c}{SR3} & \multicolumn{2}{c}{SR2} & \multicolumn{2}{c}{SR1} & \multicolumn{2}{c|}{Average} & \multicolumn{2}{c}{SN1} & \multicolumn{2}{c}{SN2} & \multicolumn{2}{c}{SN3} & \multicolumn{2}{c}{Average} \\
      & Method & AUROC$\uparrow$ & FPR95$\downarrow$ & AUROC$\uparrow$ & FPR95$\downarrow$ & AUROC$\uparrow$ & FPR95$\downarrow$ & AUROC$\uparrow$ & FPR95$\downarrow$ & AUROC$\uparrow$ & FPR95$\downarrow$ & AUROC$\uparrow$ & FPR95$\downarrow$ & AUROC$\uparrow$ & FPR95$\downarrow$ & AUROC$\uparrow$ & FPR95$\downarrow$ \\
      \midrule
      \multirow{16}{*}{\rotatebox[origin=c]{90}{\textbf{Requiring training dataset}}}& MSP\cite{paper1} & 88.1 & 67.3 & 80.6 & 84.0 & 73.7 & 80.3 & 80.8 & 77.2 & 74.3 & 82.8 & 80.0 & 78.1 & 89.7 & 52.2 & 81.3 & 71.0 \\
      & MLS\cite{vazeopen} & 89.4 & 53.8 & 83.4 & 73.1 & 76.4 & 75.3 & 83.0 & 67.4 & 72.0 & 80.8 & 83.9 & 64.1 & 89.8 & 40.5 & 81.9 & 61.8 \\
      & ODIN\cite{liang2018enhancing} & 90.2 & 47.9 & 83.3 & 71.7 & 76.3 & 76.8 & 83.3 & 65.5 & 74.2 & 79.4 & 79.4 & 71.7 & 87.8 & 41.8 & 80.5 & 64.3 \\
      & Energy\cite{liu2020energy} & 89.5 & 50.6 & 81.6 & 75.8 & 76.6 & 75.5 & 82.6 & 67.3 & 72.1 & 81.2 & 83.0 & 64.7 & 89.8 & 39.4 & 82.0 & 61.8 \\
      & GradNorm\cite{huang2021importance} & 88.5 & 50.7 & 77.4 & 75.3 & 75.2 & 76.8 & 80.4 & 67.6 & 72.1 & 81.8 & 57.7 & 88.9 & 57.8 & 79.0 & 62.6 & 83.3 \\
      & ReAct\cite{sun2021react} & 90.3 & 48.9 & 82.4 & 75.8 & 75.4 & 77.6 & 82.7 & 67.4 & 73.7 & 79.4 & 89.6 & 52.1 & 95.0 & 27.2 & 86.1 & 52.9 \\
      & NF & 88.0 & 47.7 & 80.6 & 68.2 & 75.6 & 81.4 & 81.4 & 65.8 & 81.5 & 72.5 & 71.1 & 78.0 & 91.0 & 49.6 & 81.2 & 66.7 \\
      & OE+mixup\cite{hendrycksdeep} & 72.6 & 83.5 & 72.0 & 88.5 & 62.5 & 87.8 & 69.0 & 86.6 & 72.7 & 78.9 & 80.3 & 68.8 & 87.3 & 62.2 & 80.1 & 69.9 \\
      & ARPL+CS\cite{chen2021adversarial} & - & - & - & -& -& -& - & - & 74.8 & 80.3 & 80.7 & 72.4 & 85.4 & 50.8 & 80.3 & 67.8 \\
      & Cosine proto & \textbf{91.0} & 41.0 & 82.1 & 78.2 & 77.6 & 75.6 & 83.6 & 64.9 & 80.3 & 68.3 & 88.7 & 60.8 & 91.9 & 38.0 & 86.9 & 55.7 \\
      & CE & 85.1 & 64.4 & 78.9 & 83.9 & 73.2 & 79.1 & 79.1 & 75.8 & 83.4 & 66.8 & 89.5 & 37.7 & 92.9 & 28.1 & 88.6 & 44.2 \\
      & SupCon\cite{khosla2020supervised} & - & - & - & -& -& -& - & - & 80.9 & 75.5 & 83.5 & 68.2 & 85.1 & 45.1 & 83.2 & 62.9 \\
      & SubArcface\cite{deng2020sub} & 87.1 & 61.3 & 78.9 & 76.9 & 73.7 & 81.4 & 79.9 & 73.2 & 79.0 & 81.2 & 82.9 & 60.3 & 89.1 & 32.8 & 83.7 & 58.1 \\
      & \cellcolor{gray!20}GSP(1-shot) & \cellcolor{gray!20}79.3 & \cellcolor{gray!20}61.3 & \cellcolor{gray!20}80.9 & \cellcolor{gray!20}84.1 & \cellcolor{gray!20}88.6 & \cellcolor{gray!20}51.7 & \cellcolor{gray!20}82.9 & \cellcolor{gray!20}65.7 & \cellcolor{gray!20}91.3 & \cellcolor{gray!20}44.0 & \cellcolor{gray!20}86.4 & \cellcolor{gray!20}43.1 & \cellcolor{gray!20}83.8 & \cellcolor{gray!20}91.4 & \cellcolor{gray!20}87.2 & \cellcolor{gray!20}59.5 \\
      & \cellcolor{gray!20}GSP(5-shot) & \cellcolor{gray!20}80.7 & \cellcolor{gray!20}54.8 & \cellcolor{gray!20}90.1 & \cellcolor{gray!20}47.2 & \cellcolor{gray!20}89.1 & \cellcolor{gray!20}40.2 & \cellcolor{gray!20}86.6 & \cellcolor{gray!20}47.4 & \cellcolor{gray!20}95.6 & \cellcolor{gray!20}19.2 & \cellcolor{gray!20}94.6 & \cellcolor{gray!20}22.1 & \cellcolor{gray!20}97.0 & \cellcolor{gray!20}12.5 & \cellcolor{gray!20}95.7 & \cellcolor{gray!20}17.9 \\
      & \cellcolor{gray!20}GSP(full-shot) & \cellcolor{gray!20}90.1 & \cellcolor{gray!20}\textbf{33.0} & \cellcolor{gray!20}\textbf{93.0} & \cellcolor{gray!20}\textbf{30.5} & \cellcolor{gray!20}\textbf{92.2} & \cellcolor{gray!20}\textbf{34.4} & \cellcolor{gray!20}\textbf{91.8} & \cellcolor{gray!20}\textbf{22.6} & \cellcolor{gray!20}\textbf{97.9} & \cellcolor{gray!20}\textbf{10.1} & \cellcolor{gray!20}\textbf{97.9} & \cellcolor{gray!20}\textbf{7.9} & \cellcolor{gray!20}\textbf{98.8} & \cellcolor{gray!20}\textbf{6.0} & \cellcolor{gray!20}\textbf{98.2} & \cellcolor{gray!20}\textbf{8.0} \\
      \midrule
      \multirow{6}{*}{\rotatebox[origin=c]{90}{\textbf{Zero-shot}}} & ULIP\cite{paper25} & \textbf{79.9} & 82.7 & 62.3 & 87.5 & 57.4 & 92.3 & 66.5 & 87.5 & 88.0 & 57.1 & 76.4 & 48.7 & 75.2 & 51.3 & 79.9 & \textbf{52.4} \\
      & ULIP2\cite{paper26} & 65.7 & 80.8 & 76.4 & \textbf{71.2} & 77.4 & 78.0 & 73.2 & 76.7 & 85.7 & 52.5 & \textbf{84.1} & \textbf{44.3} & 67.6 & 74.4 & 79.1 & 57.1 \\
      & DDCS\cite{paper5} & 63.3 & 90.5 & 62.4 & 80.0 & 64.4 & 96.2 & 63.4 & 88.9 & 61.1 & 86.7 & 66.8 & 86.5 & 77.2 & 76.2 & 68.4 & 83.1 \\
      & MCM\cite{ming2022delving} & 52.0 & 92.3 & 62.3 & 91.3 & 81.3 & 80.4 & 65.2 & 88.0 & 85.1 & 51.6 & 83.2 & 46.1 & 66.4 & 75.8 & 78.2 & 57.8 \\
      & NegLabel\cite{jiang2024negative} & 65.7 & 85.5 & 66.8 & 91.3 & 67.4 & 88.2 & 66.6 & 88.3 & 60.6 & 87.8 & 80.6 & 78.6 & \textbf{88.0} & \textbf{48.3} & 76.4 & 71.6 \\
    &ZLaP\cite{stojnic2024label}&70.1&81.7&67.0&84.7&67.8&83.6&68.3&83.3&88.2&52.8&72.3&66.4&77.4&90.0&79.3&69.7 \\
      & \cellcolor{gray!20}GSP & \cellcolor{gray!20}75.8 & \cellcolor{gray!20}\textbf{63.0} & \cellcolor{gray!20}\textbf{76.5} & \cellcolor{gray!20}77.6 & \cellcolor{gray!20}\textbf{83.6} & \cellcolor{gray!20}\textbf{61.6} & \cellcolor{gray!20}\textbf{78.6} & \cellcolor{gray!20}\textbf{67.4} & \cellcolor{gray!20}\textbf{90.6} & \cellcolor{gray!20}\textbf{38.9} &\cellcolor{gray!20} 70.7 & \cellcolor{gray!20}64.0 & \cellcolor{gray!20}79.7 & \cellcolor{gray!20}93.7 & \cellcolor{gray!20}\textbf{80.4} & \cellcolor{gray!20}65.5 \\
      
      \bottomrule
    \end{tabular}%
     }
     \vspace{-0.2cm}
\caption{Evaluation of OOD detection for ScanObjNN and ShapeNetCore datasets. Each column title for SR/SN indicates the chosen known class set and the rest sets of SR/SN are unknown.}
\vspace{-0.2cm}
  \label{tab:combined}
\end{table*}
\noindent\textbf{Evaluation Metrics}. We evaluate OOD detection performance using two widely adopted metrics: AUROC and FPR95. AUROC calculates the area under the receiver operating characteristic curve, with higher values indicating better performance. FPR95 measures the false positive rate at a true positive rate of 95\%, with lower values reflecting improved performance.

\subsection{Results on OOD Detection}
\noindent\textbf{Analysis of ScanObjNN Results}: From the results presented in Tab.~\ref{tab:combined}, we derive several key insights. \textbf{i)} Our proposed method demonstrates highly competitive performance under the zero-shot protocol, achieving an improvement of over 5\% in AUROC compared to ULIP2, which uses cosine distance directly as an OOD score. This substantial enhancement suggests that employing a manifold-based approach significantly enhances the differentiation between OOD samples and ID classes. \textbf{ii)} Our few-shot results are equally impressive, rivaling even fully supervised methodologies. Notably, with 5-shot training samples, our method attains an average AUROC of 86.6\%, surpassing the best fully supervised method, Cosine~proto, which scores 83.6\%, by a margin of 3\%. Furthermore, when integrating all available training samples into the graph, our approach achieves an outstanding average AUROC of 91.8\%. \textbf{iii)} A noticeable 5\% gap in average AUROC is observed between our zero-shot and 1-shot methods. This gap underscores the critical role of incorporating real in-distribution samples in effectively distinguishing OOD samples. Consequently, practitioners are advised to introduce at least a few ID samples for OOD detection, in addition to leveraging Vision-Language Models (VLMs).
\begin{table}
  \centering
  \setlength{\tabcolsep}{2pt} 
   \renewcommand{\arraystretch}{0.99} 
  \fontsize{7}{7}\selectfont
    
  \begin{tabular}{c|cc|cc|cc} 
    \hline
    \toprule
    &\multicolumn{2}{c|}{\textbf{Sydney}}& \multicolumn{2}{c|}{\textbf{nuScenes}} &\multicolumn{2}{c}{\textbf{S3DIS}} \\
     Method& AUROC$\uparrow$ & FPR95$\downarrow$ &AUROC$\uparrow$ & \multicolumn{1}{c|}{FPR95$\downarrow$} & AUROC$\uparrow$ & FPR95$\downarrow$ \\
     \midrule
    ULIP\cite{paper25}&63.4&	81.7&53.2&	\multicolumn{1}{c|}{93.3} &53.1&95.1\\
    ULIP2\cite{paper26}&66.2&	83.7&64.2&	\multicolumn{1}{c|}{93.0}&62.5&92.3	\\
    DDCS\cite{paper5}&61.8&	87.3&62.2&	\multicolumn{1}{c|}{91.2}&62.5&92.5\\
     MCM\cite{ming2022delving}&65.7&82.4&57.1&\multicolumn{1}{c|}{82.3}&65.2&88.1 \\
    NegLabel\cite{jiang2024negative}
    &58.4&92.7&69.7&\multicolumn{1}{c|}{81.6}&55.4&\textbf{85.4}\\
    ZLaP\cite{stojnic2024label}&58.5&80.6&70.0&87.9&71.3&88.4     \\
    \rowcolor{gray!20} GSP&\textbf{72.5}&	\textbf{68.9}&\textbf{80.8}&	\multicolumn{1}{c|}{\textbf{80.7}}&\textbf{77.6}&94.0\\
   
    \bottomrule
  \end{tabular}
  \vspace{-0.2cm}
  \caption{Zero-shot OOD detection results on the Sydney Urban Objects dataset, nuScenes dataset and S3DIS dataset.}
  \vspace{-0.2cm}
  \label{tab:sydneydataset}
\end{table}


\noindent\textbf{Analysis of ShapeNetCore Results}: The results for ShapeNetCore, as shown in Tab.~\ref{tab:combined}, lead us to several observations. \textbf{i)} Unlike the findings from the ScanObjNN dataset, the performance gap between the leading baseline, ULIP, and our method under the zero-shot protocol is relatively narrow, approximately $0.5\%$. This modest improvement can be attributed to the substantial overlap between the downstream and pre-training datasets used in Vision-Language Models (VLMs). Specifically, ULIP/ULIP2 are pre-trained on ShapeNet, which shares significant commonalities with ShapeNetCore, thereby impacting OOD detection capabilities. Despite this, the slight enhancement achieved by our method underscores the utility of leveraging data manifolds to improve the differentiation of OOD samples. \textbf{ii)} A marked enhancement in performance is observed when few-shot samples are available. For instance, introducing 1-shot training samples results in nearly a 7\% increase in average AUROC, while 5-shot samples yield an impressive 15\% improvement. This highlights the critical role of few-shot learning in enhancing model performance.

\noindent\textbf{Analysis of Additional Real-World Datasets}: The zero-shot OOD detection results for the Sydney Urban Objects dataset~\cite{de2013unsupervised} and nuScenes Dataset~\cite{nuScenes}, presented in Tab.~\ref{tab:sydneydataset}, reinforce our previous conclusions. Our method consistently outperforms all competing approaches in zero-shot OOD detection, demonstrating the efficacy of utilizing testing data manifolds, particularly when there is a substantial shift between the downstream task and the pre-training data. This finding emphasizes the robustness of our approach in adapting to significant distributional shifts in real-world applications.
 For S3DIS~\cite{S3DIS}, GSP achieves the highest AUROC, outperforming all baselines by a significant margin (e.g., +12.4\% over MCM). This suggests superior ability to distinguish between classes across all decision thresholds. However, GSP’s FPR95 ($94.0\%$) is slightly higher than ULIP ($95.1\%$), indicating that while GSP excels in overall classification, it faces challenges in suppressing false positives when operating at a 95\% true positive rate.

\noindent\textbf{Further Insights Synergizing All Results}: By comparing the outcomes from the ScanObjNN, Sydney Urban Objects, S3DIS, nuScenes and ShapeNetCore datasets, we derive several overarching insights. \textbf{i)} There is a notable correlation between the distribution shift from pre-training to downstream datasets and the efficacy of manifold exploitation for OOD detection. This can be attributed to several factors. Firstly, when pre-training is conducted on datasets that closely resemble the downstream datasets, the contrastive training loss effectively aligns text prompts with visual features, resulting in a well-defined separation between ID and OOD samples. Conversely, in scenarios where the distribution shift is more substantial, VLMs may struggle to accurately quantify the distance between ID text prompts and visual samples. In such cases, leveraging the testing data manifold becomes essential to enhance the calibration of prompt-to-visual feature distances, thereby improving OOD detection performance.

\noindent\textbf{}

\subsection{Ablation Study}


We begin by analyzing the performance of three Out-of-Distribution (OOD) detection methods: Cosine Similarity, Shortest Path, and Label Propagation. As shown in Tab.~\ref{tab:ablation}, we employ the ULIP2 backbone for extracting text and point cloud features and evaluate these methods for OOD scoring. The approach of directly computing the maximal normalized cosine similarity between text and point cloud features yields a baseline performance. Substituting cosine similarity with the shortest path metric results in a performance increment, albeit modest. It is noteworthy that the shortest path is computed using Dijkstra's algorithm, and the reciprocal of manifold distance serves as the OOD score. Lastly, we introduce a label propagation method to determine the OOD score. The experimental outcomes in Tab.~\ref{tab:ablation} demonstrate that integrating graph-based information via label propagation significantly enhances the model's ability to differentiate between In-Distribution (ID) and OOD categories. This suggests that graph-based methods provide a more nuanced understanding of the feature space, leading to improved OOD detection capabilities.

\begin{table}
  
  \centering
  \setlength{\tabcolsep}{3pt} 
   \renewcommand{\arraystretch}{0.9} 
  \fontsize{8}{9}\selectfont
  \begin{tabular}{c|cc|cc} 
    \hline
    \toprule
     OOD Det. Mthd. &	Prompt Clust.&	 Neg. Prompt &  AUROC$\uparrow$ & FPR95$\downarrow$\\
     \midrule
     Cosine Dist.& -	&	- &73.2 &	76.7\\
     Manifold Dist.&	- &	- &74.1	&73.0\\
     Score Prop.&	- &	- &76.4&72.6\\
     Score Prop.&	\checkmark&&77.2&73.6\\
     Score Prop.&	-	&\checkmark&77.3&71.4\\
     Score Prop.&	\checkmark& \checkmark&\textbf{78.6}&\textbf{67.4}\\
    \bottomrule
  \end{tabular}
  \vspace{-0.2cm}
\caption{Ablation study on ScanObjNN dataset.}
\vspace{-0.2cm}
  \label{tab:ablation}
\end{table}



\subsection{Further Analysis}

\begin{figure}[t]
  \centering
  \includegraphics[width=1.00\linewidth]{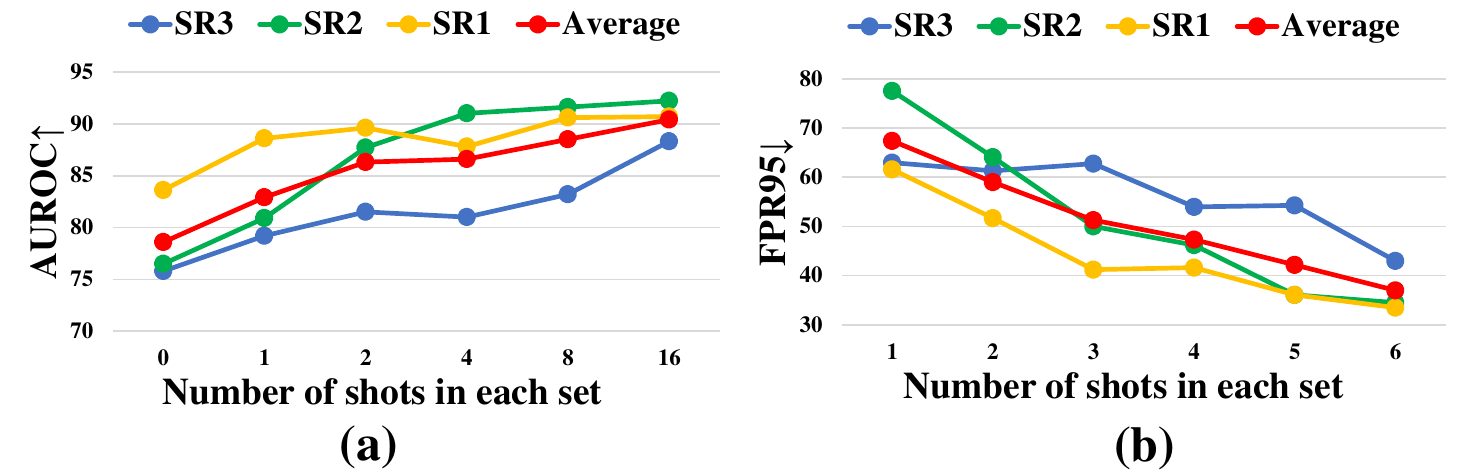}
  \vspace{-0.5cm}
  \caption{Few-shot OOD detection on the ScanObjNN dataset. (a) Results on AUROC metric. (b) Results on FPR95 metric.}
  \label{figureshots}
\end{figure}

\noindent\textbf{Few-shot OOD Detection}: To evaluate the performance of label propagation in few-shot OOD detection, we incorporated a subset of training data from the 3DOS training dataset, treating them as in-distribution categories. The outcomes, presented in Fig.~\ref{figureshots}, show the effects of varying the number of training samples used in label propagation, with all samples randomly selected from the training dataset. The results indicate that as the number of training samples increases, OOD detection performance improves and eventually stabilizes. 

\noindent\textbf{Qualitative Analysis}: To validate our method, we visualize sample predictions and scores from baseline method and our full GSP. As shown in Fig.~\ref{figure9}, for easy-to-predict samples from the Real-world dataset, conventional distance-based methods perform well. However, in challenging cases (e.g., occlusions or missing parts), they often fail. In contrast, our method utilizes spatially proximate samples in feature space for score propagation, ensuring accurate OOD detection even in difficult scenarios. These qualitative results confirm the effectiveness and robustness of our approach.

\begin{figure*}[!htb]
  \centering
  \includegraphics[width=1\linewidth]{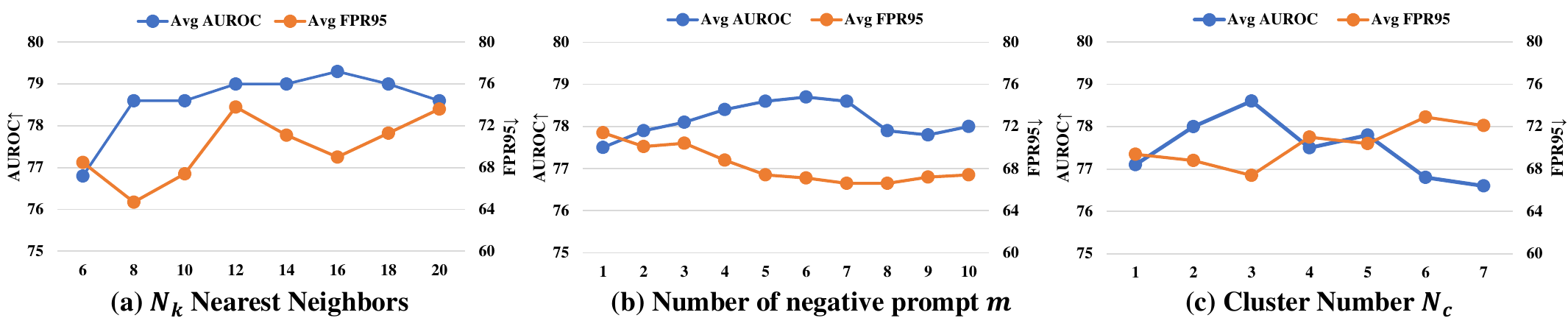}

  \vspace{-0.3cm}
  \caption{Results from ScanObjNN Benchmark Experiments: (a) Evaluation of prompt clustering with different K-means cluster numbers. (b) Impact of varying negative prompts on label propagation. (c) Results of different number of K Neaerst Neighors in KNN.}
  \label{fig:prompt_clust}
\end{figure*}

\begin{figure}[!hbt]
  \centering
  \includegraphics[width=1\linewidth]{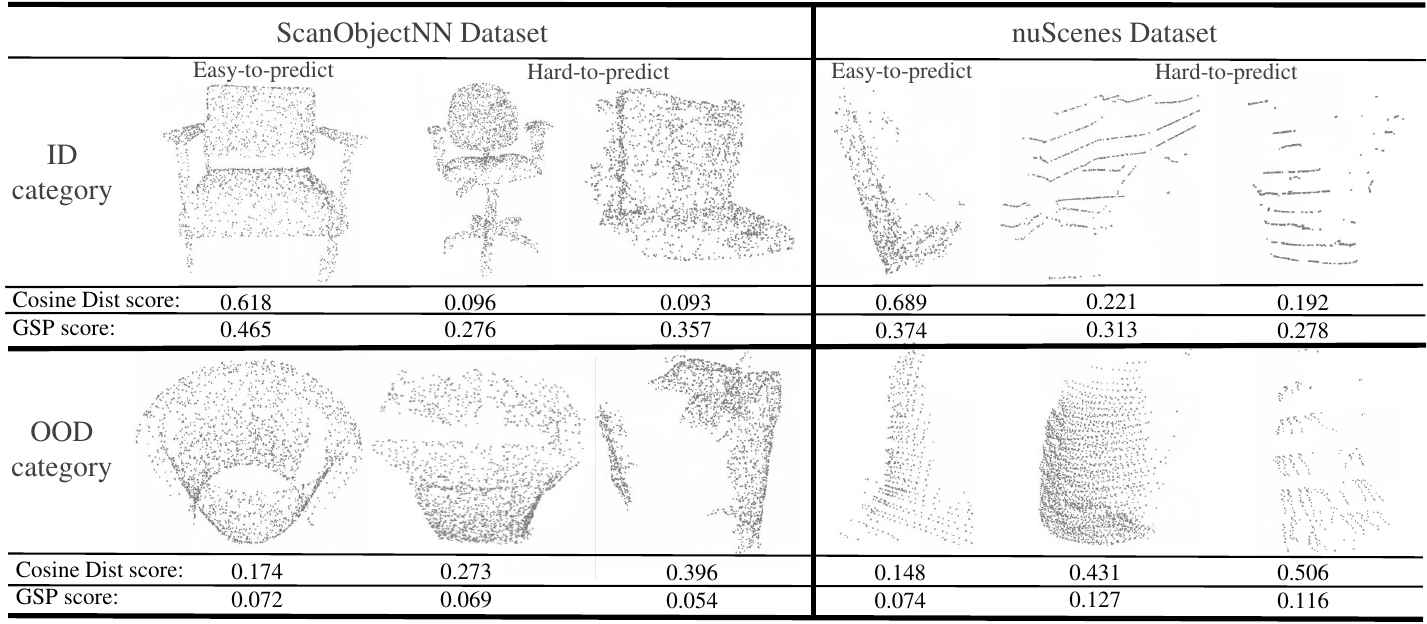}
  \vspace{-0.6cm}
  \caption{Visualization of some samples for ScanObjectNN and nuScenes Dataset with two methods predict scores. All scores are normalized to 0-1. }
  \label{figure9}
  \vspace{-0.2cm}
\end{figure}

\noindent\textbf{Impact of Prompt Clustering}: To illustrate the benefits of prompt expansion via K-means clustering, we examined how varying the number of prompts affects OOD (Out-of-Distribution) detection. As depicted in Fig.~\ref{fig:prompt_clust} (a), we experimented with different values of \( N_c \) in the K-means algorithm, where each \( N_c \) value corresponds to a distinct number of clusters, and consequently, a different number of prompts. An increase in \( N_c \) results in a greater number of distinct prompts.

The results presented in the figure reveal that the quantity of prompts considerably impacts OOD detection performance. Specifically, when \( N_c = 6 \), the AUROC surpasses 78.7\%. However, further increasing \( N_c \) can negatively affect the spatial alignment between text features and point cloud features. This phenomenon is likely due to the fact that typical prompts in Vision-Language Models (VLMs) are generally standardized, with only occasional outliers. Excessive clustering may isolate these outliers, leading to a decline in performance.

In conventional VLM approaches, the mean text feature vector is typically calculated for each category (equivalent to setting \( N_c = 1 \)), achieving an average AUROC of less than 78\%. The findings in Fig.~\ref{fig:prompt_clust} (c) suggest that selecting an appropriate \( N_c \) enhances the model's ability to differentiate OOD categories, underscoring the advantages of prompt clustering.

\noindent\textbf{Effectiveness of Negative Prompting via Self-Training}: We conducted experiments to evaluate the effectiveness of the proposed negative prompting. Using the ShapeNetCore dataset, we explored the impact of self-training for negative prompting by varying the percentage of samples tagged with pseudo labels, denoted as \( m\% \) as introduced in Section~\ref{sect:OODLP}. 
In Fig.~\ref{fig:prompt_clust} (b), We observe a performance increase from \( m = 1 \) up to \( m = 6 \), indicating that incorporating negative prompts through self-training is an effective strategy to leverage the manifold for score propagation. However, performance declines beyond \( m = 6 \), likely due to the introduction of incorrect pseudo-negative prompts. This highlights the importance of mitigating the effects of confirmation bias~\cite{arazo2020pseudo} to prevent the model from being adversely affected.

\noindent\textbf{Impact of K-Nearest-Neighbor.} We compare performance with different values of \( N_k \) Nearest Neighbors. As we can see in Fig \ref{fig:prompt_clust} (a), the results demonstrate that as the number of \( N_k \) values increases, the OOD detection
performance is tend to be stable. While \( N_k \) = 16 achieve the best performance and Less than 8 or lager than 18 get the low results. 

\noindent\textbf{Computational Cost}: We profiled each step and test the inference time of GSP using a full GPU implementation in Tab.~\ref{tab:testtime}. Experiments are carried out on a single RTX3090 GPU and a i7-12700k CPU. As observed, the graph-based operations (Build Graph + Score Propagation) account for only around 2\% of the total inference time. The additional computation cost introduced by graph operations is negligible compared with point cloud feature extraction.



   

     

\begin{table}
  \centering
  \setlength{\tabcolsep}{0.5pt} 
   \renewcommand{\arraystretch}{0.9} 
  \fontsize{6}{6}\selectfont
  \resizebox{0.5\textwidth}{!}{
  \begin{tabular}{c|cccccc|cc} 
    \hline
    \toprule
    
   &Modelload + &Text Feature&Prompt&Feature&Build &Score& Total\\
&Datasetload&Extract&Clusting&Extract&Graph&Propagation&time \\
   \midrule
    SR&1552ms&2212ms&1128ms&2780ms&106ms&8ms&7786ms	\\
    SN&1641ms&2459ms&1657ms&25031ms&671ms&34ms & 31493ms 	\\
    \bottomrule
  \end{tabular}}
  \vspace{-0.35cm}
 \caption{\footnotesize{The inference time of the GSP on SR and SN.}}
 \vspace{-0.4cm}
  \label{tab:testtime}
\end{table}



\section{Conclusion}
In this work, we presented a novel training-free approach to 3D point cloud OOD detection that effectively leverages Vision-Language Models through Graph Score Propagation. By constructing a graph that incorporates both positive prototypes and pseudo-negative samples via self-training, our method captures the underlying structure of the testing data manifold. This approach allows for the propagation of informative OOD scores, thus surpassing existing distance-based techniques. Empirical results across multiple 3D point cloud datasets demonstrate the robustness and adaptability of our framework, particularly in zero- and few-shot settings. The GSP method addresses key limitations in prior OOD approaches by reducing dependency on extensive in-distribution training data while achieving high accuracy. Future work could explore further enhancements through dynamic graph updates, enabling real-time OOD detection in continuously evolving environments.

{
    \small
    \bibliographystyle{ieeenat_fullname}
    \bibliography{main}
}

\clearpage
\setcounter{page}{1}
\maketitlesupplementary
\section*{A. Details of Datasets}

\noindent\textbf{Partition of dataset}:
We begin by following the 3DOS methodology to partition the ShapeNetCore~(Synthetic benchmark) and ScanObjectNN~(Real benchmark) datasets. For the Real benchmark, synthetic point clouds from ModelNet40 are used for training, while testing is conducted on real-world point clouds from ScanObjectNN. For the Synthetic benchmark, the ShapeNetCore dataset is divided into three non-overlapping (i.e., semantically distinct) category sets, each containing 18 categories. The specific categories for the SR/SN sets are detailed in Tab.~\ref{tab:3DOSdataset}, with illustrative examples shown in Fig.~\ref{supp_figure3}.

The Sydney Urban Objects Dataset provides high-resolution point clouds that capture detailed geometric information about various urban objects, such as buildings, vehicles, trees, and street furniture. To demonstrate the application of out-of-distribution (OOD) methods in real-world scenarios, we split the dataset into two parts: movable objects and non-movable objects. Movable objects are treated as in-distribution (ID) classes, while non-movable objects are considered OOD classes.

The Stanford Large-Scale 3D Indoor Spaces (S3DIS) dataset is a comprehensive benchmark for 3D indoor scene understanding. To ensure sufficient geometric fidelity, we constructed the test dataset by retaining instances with over 2,048 raw points, resulting in 8,931 high-density instances.

The nuScenes dataset is a multimodal autonomous driving resource featuring 1,000 urban scenes captured via LiDAR, radar, and cameras, with 3D annotations. Given the scale of the nuScenes dataset, we utilized its trainval.mini subset to balance data diversity and computational feasibility. Point clouds with fewer than 200 raw points were discarded to ensure reliable geometric representation, resulting in a curated collection of 2205 high-resolution LiDAR frames to constructing testing dataset.

Some ID classes and OOD classes of Sydney Urban Objects, S3DIS and nuScenes dataset are shown in Fig.~\ref{supp_figure4}. The category classification is provided in Tab.~\ref{tab:sydneysplit}.

\begin{figure}[t]
  \centering
  \includegraphics[width=0.95\linewidth]{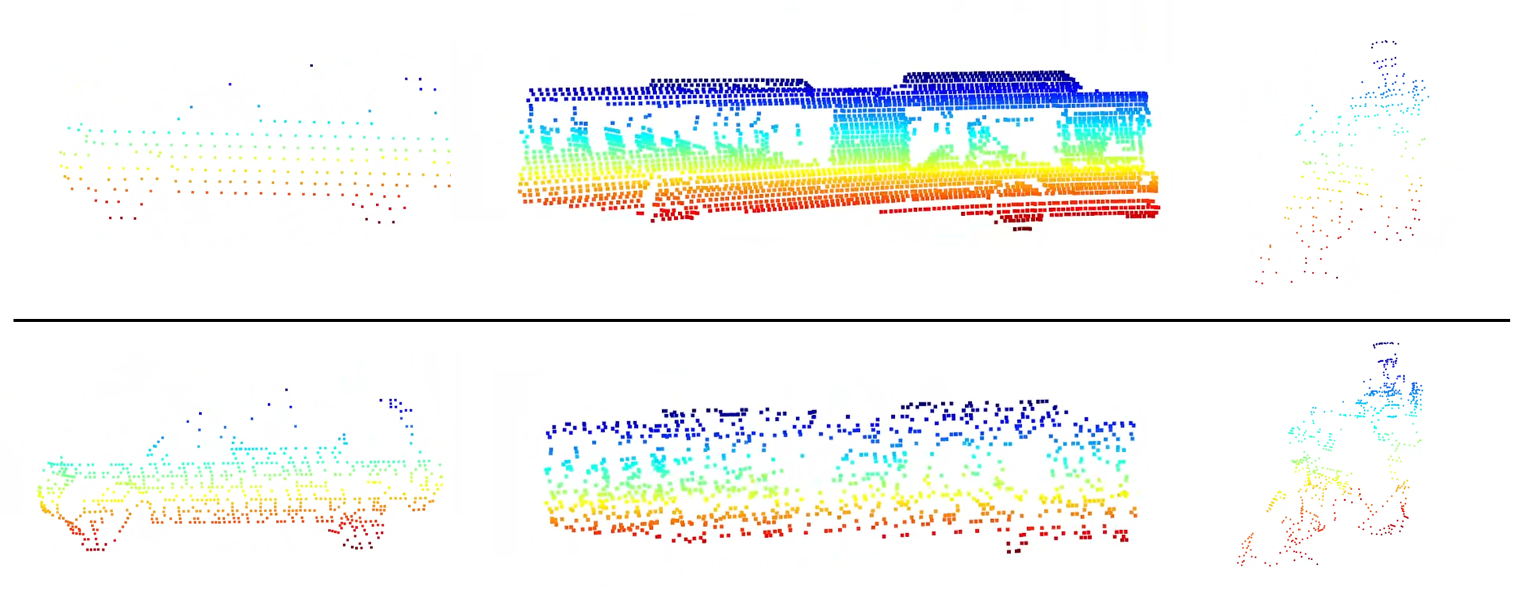}
  \vspace{-0.2cm}
  \caption{Illustration of point cloud interpolation results. We uniformly subsample or interpolate each sample to 2048 points.}
  \label{supp_Figure2}
  \vspace{-0.3cm}
\end{figure}

\noindent\textbf{Dataset Preprocessing}: 
For the Sydney Urban Objects and nuScenes Dataset, the number of points in each frame of the point cloud varies, as the data is cropped from real-world scenes. To ensure consistency, we randomly sample each point cloud to standardize it to 1024 points per point cloud. Figure~\ref{supp_Figure2} illustrates examples of the original point clouds alongside their normalized counterparts. It is evident that certain categories, such as "4wd" and "biker," contain fewer points in the original point cloud, whereas others, like "bus," have significantly more points. 

To address these discrepancies, we implement a data augmentation strategy. For categories with fewer points, we generate synthetic points by interpolating between each original point and its nearest neighbors. Specifically, for each point, a random neighbor is selected, and a specified number of new points are created through linear interpolation. The interpolation coefficient is adjusted to distribute the new points evenly between the original point and its neighbor. For categories with excess points, we apply subsampling to reduce the point count. This approach ensures that all point clouds are consistently normalized to 1024 points before being input into the model.


\noindent\textbf{Prompt templates}: We follow the ULIP framework to construct prompts by applying each class name to 64 predefined templates, generating a comprehensive set of text descriptions for Prompt Clustering. The full list of prompt templates is provided in Table~\ref{prompt}. We empirically find that including the template "a photo of {}" to be beneficial for 3D point cloud OOD detection. This is due to the fact that ULIP is built upon CLIP by aligning point cloud feature to text and image (rendered from 3D point cloud) features. Thus the text prompt of ``a photo of {}'' is a legacy from CLIP.

\begin{figure*}[t] 
  \centering
  \includegraphics[width=0.99\linewidth]{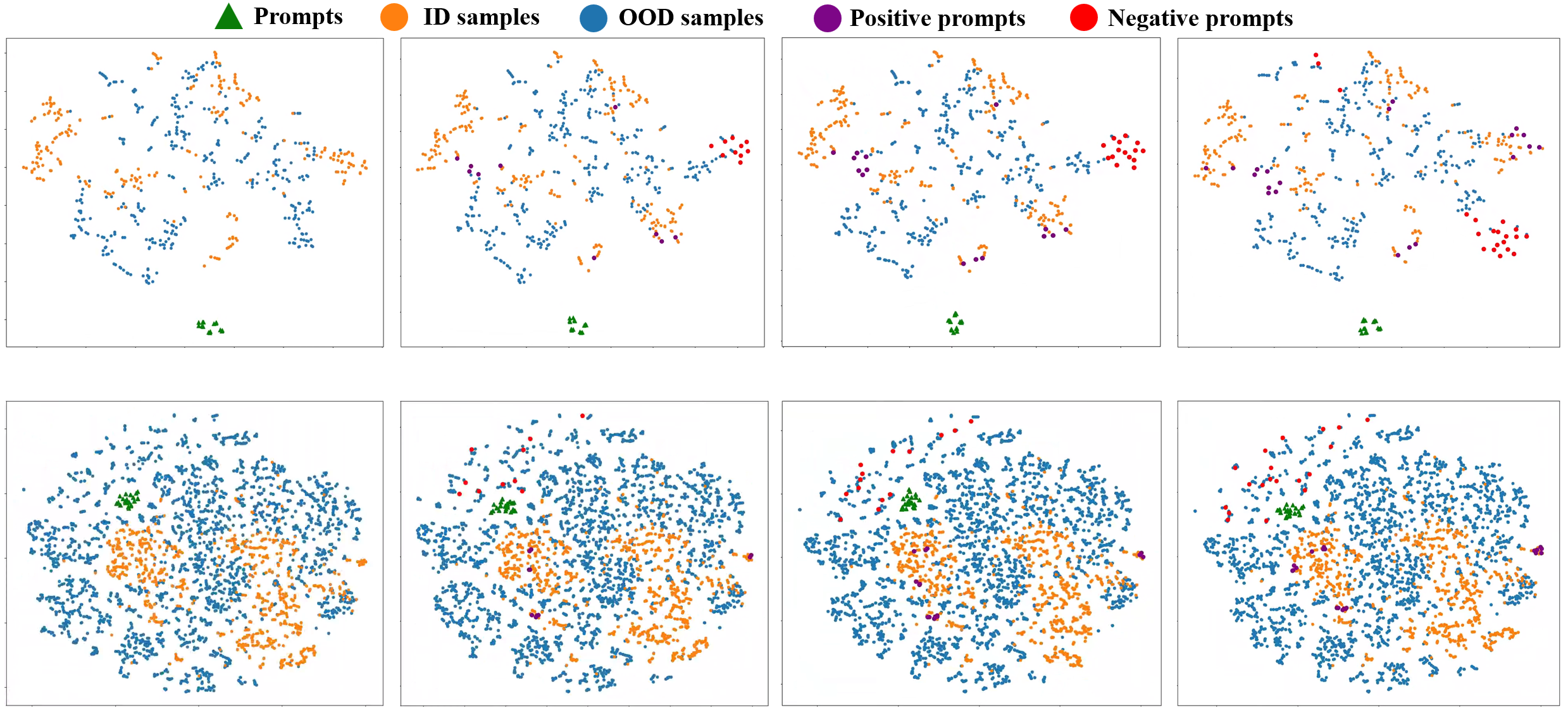}
  \vspace{-0.2cm}
  \caption{T-SNE Visualization on the ScanObjNN and ShapNetCore Dataset for self-training with different number of negative prompts. The
first row displays the visualization on the ScanObjNN, and the second row for ShapNetCore Dataset. }
  \vspace{-0.3cm}
  \label{supp_figure1}
\end{figure*}

\section*{B. Additional Studies}


\noindent\textbf{Study of Self-Training}: To investigate the self-training process, we visualized the feature distributions in Fig.~\ref{supp_figure1}. Different rows correspond to various experimental datasets, while different columns represent varying numbers of negative prompts. The first column shows the initialization of all features, and the subsequent columns illustrate the feature distributions when the number of negative prompts is set to 10, 15, and 20, respectively. From this figure, we observe that positive prompts tend to cluster closer to ID samples, while negative prompts align more closely with OOD samples. This distinction helps assign higher scores to ID samples and lower scores to OOD samples during the label propagation stage. These findings highlight the critical role of generating negative prompts through self-training in enhancing the effectiveness of the process.

\noindent\textbf{VLM backbone}: ULIP was chosen as the default VLM backbone due to its superior performance. However, the proposed GSP method is agnostic to the VLM backbone, as demonstrated by additional evaluations using PointClip~V2 as the backbone in Tab.\ref{tab:pointclipv2}. GSP consistently shows superior performance with PointClip~V2 as well.

\begin{table*}
  \centering
  \setlength{\tabcolsep}{2pt} 
  \renewcommand{\arraystretch}{0.4} 
  \fontsize{7}{7}\selectfont
  \resizebox{\textwidth}{!}{
    \begin{tabular}{c|cc|cc|cc|cc|cc|cc|cc|cc}
    
      \hline 
      \toprule
       & \multicolumn{2}{c}{SR3} & \multicolumn{2}{c}{SR2} & \multicolumn{2}{c}{SR1} & \multicolumn{2}{c|}{Average} & \multicolumn{2}{c}{MN1} & \multicolumn{2}{c}{MN2} & \multicolumn{2}{c}{MN3} & \multicolumn{2}{c}{Average} \\
      Method & AUROC$\uparrow$ & FPR95$\downarrow$ & AUROC$\uparrow$ & FPR95$\downarrow$ & AUROC$\uparrow$ & FPR95$\downarrow$ & AUROC$\uparrow$ & FPR95$\downarrow$ & AUROC$\uparrow$ & FPR95$\downarrow$ & AUROC$\uparrow$ & FPR95$\downarrow$ & AUROC$\uparrow$ & FPR95$\downarrow$ & AUROC$\uparrow$ & FPR95$\downarrow$ \\
      \midrule
      Cosine Dist&36.5&96.1&62.2&92.9&63.3 &83.2  &54.0 &90.7  &68.2&83.3&69.3&79.7&71.9&75.0	& 69.8&79.3\\
    GSP(Ours)&48.5&81.4& 63.1&93.4&67.6&81.2& 59.7&85.3 &71.8&68.9&73.3&64.9&69.2&74.1&71.4&69.3\\
 \bottomrule
      
    \end{tabular}%
     }
     \vspace{-0.28cm}
\caption{The results of distance-based and GSP on Modelnet40 and ScanObjectNN dataset with PointClip~V2 as VLM backbone.}
\vspace{-0.3cm}
  \label{tab:pointclipv2}
\end{table*}

\begin{figure}[t]
  \centering
  \includegraphics[width=0.8\linewidth]{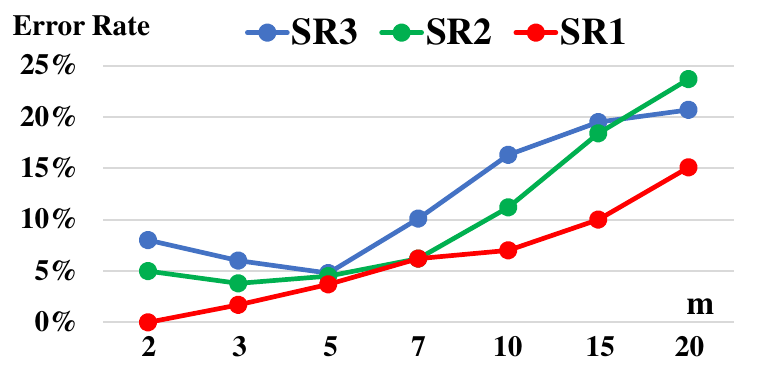}
  \vspace{-0.2cm}
  \caption{Error rates of pseudo prompts with different filtering ratio $m$.}
  \vspace{-0.55cm}
  \label{fig:ERROR}
\end{figure}

\begin{figure}[t]
  \centering
  \includegraphics[width=0.8\linewidth]{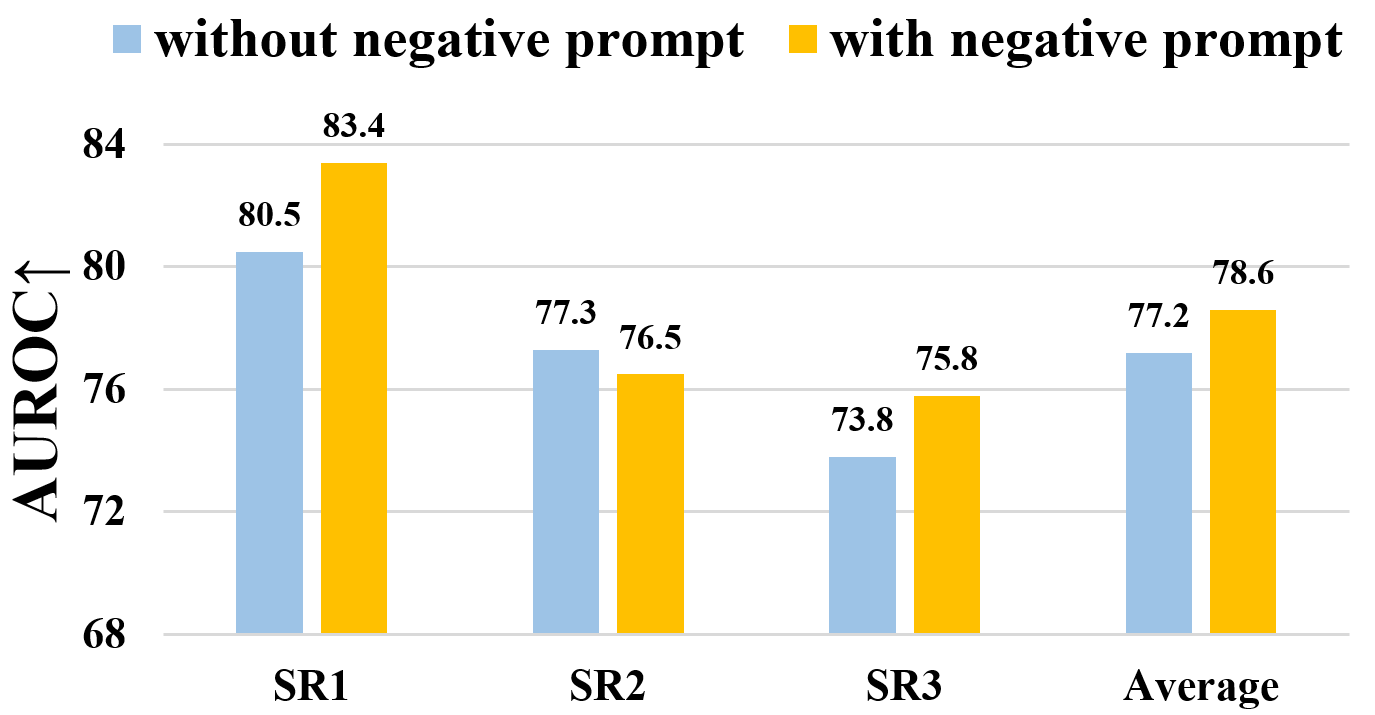}
  \vspace{-0.1cm}
  \caption{Ablation Study on the Impact of Negative Prompts: Blue Bars Indicate Results Without Negative Prompts, While Yellow Bars Indicate Results With Negative Prompts. All Experiments Conducted on the ScanObjNN Dataset.}
  \label{figure5}
  \vspace{-0.25cm}
\end{figure}
\noindent\textbf{Incorrect pseudo prompts}: Self-training is known to be sensitive to incorrect pseudo labels. To further analyze the impact of pseudo prompt selection, we vary $m$ from 2 to 20 and examine the error rate on selected pseudo prompts (Fig.~\ref{fig:ERROR}). As $m$ decreases (stricter selection), the error rate on pseudo prompts reduces. On average, $m=5$ proves to be an effective hyperparameter for selecting reliable pseudo prompts.

\noindent\textbf{Study of negative prompt}: The ablation study in Fig~\ref{figure5} evaluates the impact of negative prompts on ScanObjectNN Dataset, the results reveal a consistent performance gap between models with and without negative prompts.


\begin{table*}
  \centering
  \setlength{\tabcolsep}{9pt} 
   \renewcommand{\arraystretch}{0.9} 
  \fontsize{8}{9}\selectfont
    
  \begin{tabular}{c|c|c} 
    \hline
    \toprule
    \multicolumn{3}{c}{The Real benchmark(ScanobjNN)}\\
    \midrule
     SR1& SR2&SR3  \\
     \midrule
     chair, shelf, door, sink, sofa&bed, toilet, desk, table, display & bag, bin, box, pillow,
 cabinet \\
     \midrule
    \multicolumn{3}{c}{The Synthetic benchmark(ShapeNetCore)}\\
    \midrule
    SN1&SN2&SN3\\
    \midrule
    mug, lamp, bed, washer, loudspeaker& earphone, knife, chair, pillow, table, 
  &  can, microwave, skateboard, faucet, train\\
 telephone, dishwasher, camera, birdhouse, jar& 
 mailbox, basket, file cabinet, cabinet, sofa &  pistol, helmet, watercraft, airplane, bottle\\ bowl, bookshelf, stove, bench, display&flowerpot, microphone, tower, bag, bathtub& rocket,
 rifle, remote, car, bus\\, keyboard, clock, piano&laptop, printer, trash bin& guitar, cap, motorbike\\

    \bottomrule
  \end{tabular}
  \vspace{-0.2cm}
  \caption{Classification of Categories in the ShapeNetCore and ScanobjNN Dataset.}
  \vspace{-0.2cm}
  \label{tab:3DOSdataset}
\end{table*}

\begin{table*}
  \centering
  \setlength{\tabcolsep}{9pt} 
   \renewcommand{\arraystretch}{0.9} 
  \fontsize{8}{9}\selectfont
    
  \begin{tabular}{c|c|c} 
    \hline
    \toprule
     Dataset&in-distribution categories&out-of-distribution categories\\
    \midrule
    The Sydney&bus, car, cyclist, excavator, pedestrian, scooter, & trash, tree, trunk, umbrella, ute, pillar, pole, post, building,
\\
Urban Objects &4wd, bicycle, biker, trailer, truck, van& 
   bench, ticket machine, traffic lights, traffic sign, vegetation\\
   
   \midrule
   S3DIS&window, door, table, chair, clutter, sofa, bookcase & floor, ceiling, wall, beam, board, column\\
   
   \midrule
   nuScenes&pedestrain, car, motorcycle & barrier, truck, bus, traffic\_cone, construction\_vehicle, trailer\\
    \bottomrule
  \end{tabular}
  \vspace{-0.2cm}
  \caption{Classification of Categories in the Sydney Urban Objects Dataset, S3DIS dataset and nuScenes dataset.}
  \vspace{-0.2cm}
  \label{tab:sydneysplit}
\end{table*}

\begin{table*}
  \centering
  \setlength{\tabcolsep}{7pt} 
   \renewcommand{\arraystretch}{0.9} 
  \fontsize{8}{9}\selectfont
    
  \begin{tabular}{llll} 
    \hline
    \toprule
    \multicolumn{4}{c}{Prompt templates}\\
    \midrule
    “a point cloud model of \{\}.”&  “There is a \{\} in the scene.” &    “There is the \{\} in the scene.” &  “a photo of a \{\} in the scene.” \\“a photo of the \{\} in the scene.”&“a photo of one \{\} in the scene.” & “itap of a \{\}.”& “itap of my \{\}.” \\“itap of the \{\}.”&  “a photo of a \{\}.”& “a photo of my \{\}.” &“a photo of the \{\}.”\\“a photo of one \{\}.”&  “a photo of many \{\}.”& “a good photo of a \{\}.”& "a good photo of the \{\}."\\“a bad photo of a \{\}.” &“a bad photo of the \{\}.”&“a photo of a nice \{\}.”&   “a photo of the nice \{\}.” \\“a photo of a cool \{\}.” & “a photo of the cool \{\}.”&“a photo of a weird \{\}.”&“a photo of the weird \{\}.”\\“a photo of a small \{\}.” &“a photo of the small \{\}.”& “a photo of a large \{\}.”&“a photo of the large \{\}.”\\“a photo of a clean \{\}.”&“a photo of the clean \{\}.”&“a photo of a dirty \{\}.”&“a photo of the dirty \{\}.”\\“a bright photo of a \{\}.”&“a bright photo of the \{\}.”&“a dark photo of a \{\}.” & “a dark photo of the \{\}.”\\“a photo of a hard to see \{\}.”&“a photo of the hard to see \{\}.”&“a low resolution photo of a \{\}.” &“a cropped photo of a \{\}.”\\“a low resolution photo of the \{\}.”&    “a cropped photo of the \{\}.”&“a close-up photo of a \{\}.”&“a close-up photo of the \{\}.”\\“a jpeg corrupted photo of a \{\}.”& “a jpeg corrupted photo of the \{\}.”&“a blurry photo of a \{\}.”&   “a blurry photo of the \{\}.”\\“a pixelated photo of a \{\}.”&     “a pixelated photo of the \{\}.”&“a black and white photo of the \{\}.”&“a plastic \{\}.”\\“a black and white photo of a \{\}”&     “the plastic \{\}.”&“a toy \{\}.” & “the toy \{\}.”\\“a plushie \{\}.”&“the plushie \{\}.”& “a cartoon \{\}.”& “the cartoon \{\}.”\\
    “an embroidered \{\}.”& “the embroidered \{\}.” & “a painting of the \{\}.” & "a painting of a \{\}."\\
    
    \bottomrule
  \end{tabular}
  \vspace{-0.2cm}
  \caption{All prompt templates for GSP.}
  \vspace{-0.2cm}
  \label{prompt}
\end{table*}

\section*{C. Broader Impact and Limitations}

\subsection*{C.1 Broader Impact}

The proposed method could improve the efficacy of generalizing pre-trained 3D VLM for real-world OOD detection tasks. Adopting the techniques could benefit autonomous driving and robotics, potentially improving the safety. Potential risks include failing to differentiate OOD from ID may lead to collision and fatal consequences.

\subsection*{C.2 Potential Limitations}

The method requires seeing a substantial amount of testing data so that a graph can be built and inference can benefit from the manifold information. If testing data arrives in a stream, the proposed method could be implemented in a batch mode or incrementally build a graph for inference.

\begin{figure*}[!htb]
  \centering
  \includegraphics[width=0.99\linewidth]{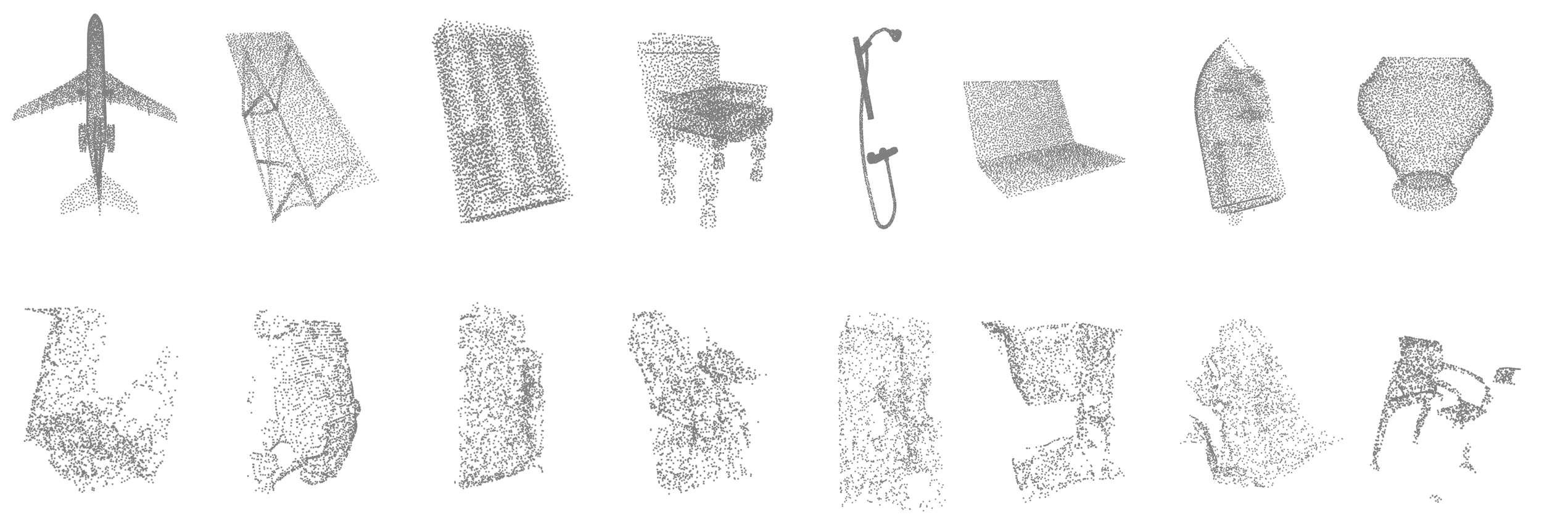}
\vspace{-0.2cm}
  \caption{Visualization of some point clouds from the ScanObjNN and ShapeNetCore Dataset. The first row displays the visualization on the  ShapNetCore , and the second row for ScanObjNN Dataset. }
  \vspace{-0.3cm}
  \label{supp_figure3}
\end{figure*}

\begin{figure*}[t] 
  \centering
  \includegraphics[width=0.99\linewidth]{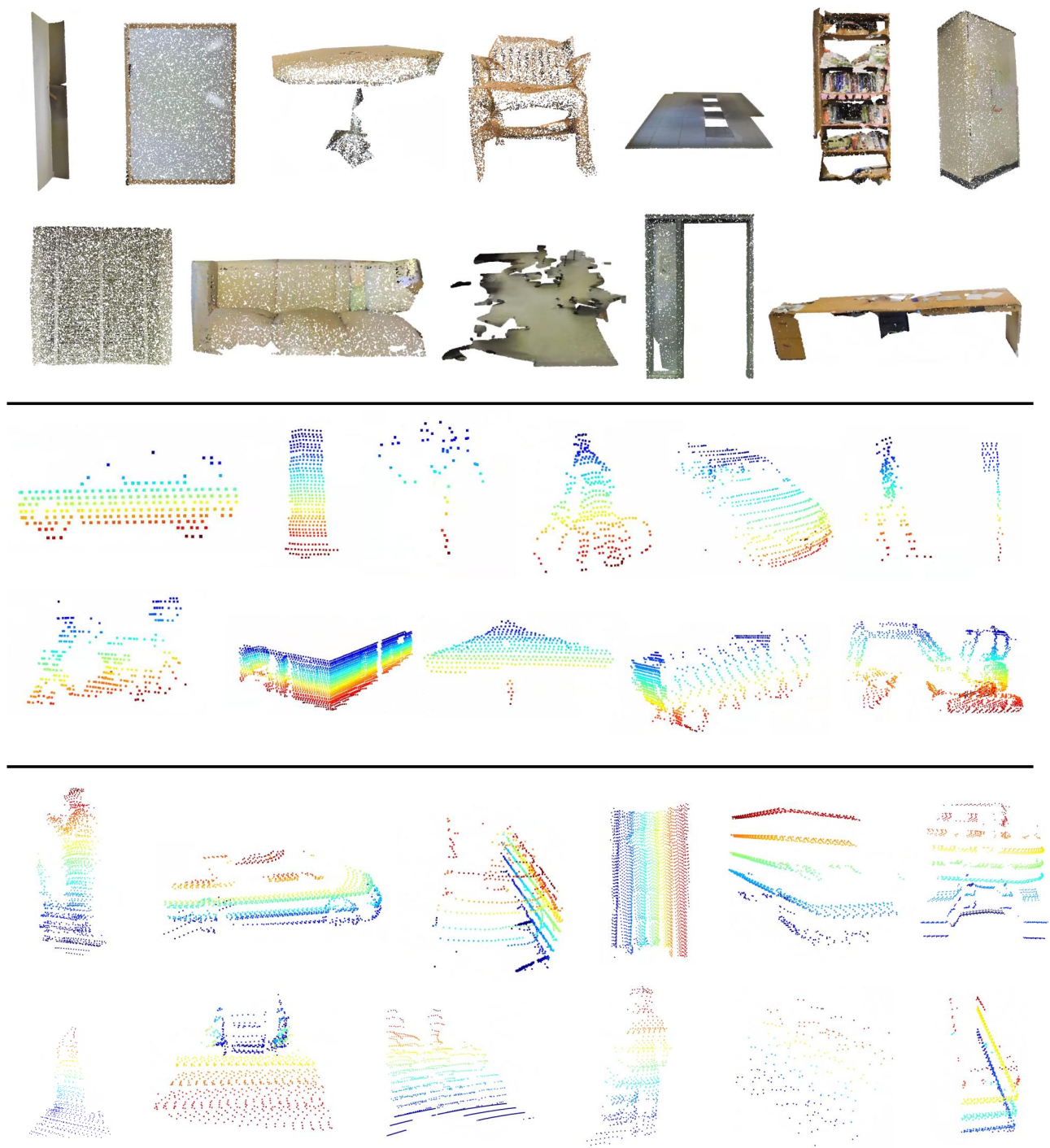}
  \vspace{-0.2cm}
  \caption{Visualization of some point clouds from the S3DIS, Sydney Urban Objects and nuScenes Dataset. The first row displays the visualization on the S3DIS, the second row for Sydney Urban Objects, and the third row for nuScenes Dataset.}
  \vspace{-0.3cm}
  \label{supp_figure4}
\end{figure*}

\end{document}